\documentclass[journal]{IEEEtran}

\usepackage{times,amsmath,epsfig,amssymb,graphicx,amsfonts,pxfonts,txfonts,float}
\usepackage{subfigure}
\usepackage{psfrag}
\usepackage{ifpdf}
\usepackage{cite}

\begin{document}

\title{Compensating Interpolation Distortion by New Optimized Modular Method}

\author{Mohammad~Tofighi,~\IEEEmembership{Member,~IEEE,}
        Ali~Ayremlou,~\IEEEmembership{Member,~IEEE,}
        and~Farokh Marvasti,~\IEEEmembership{Member,~IEEE}%
\thanks{Manuscript received ..., 2012 (the paper is ready and will be submitted soon).}%
\thanks{M. Tofighi  is with Department of Electrical Engineering, Urmia University, Urmia, Iran e-mail: (mo.tofighi@gmail.com).}%
\thanks{A. Ayremlou and F. Marvasti are with Advanced Communication Research Institute (ACRI), Department of Electrical Engineering, Sharif University of Technology, Tehran, Iran e-mail: (a\_ayremlou@ee.sharif.edu; marvasti@sharif.edu).}}

\markboth{to be submitted}%
{Shell \MakeLowercase{\textit{et al.}}: to be submitted}

\maketitle

\begin{abstract}
A modular method was suggested before to recover a band limited signal from the sample and hold and linearly interpolated (or, in general, an nth-order-hold) version of the regular samples. In this paper a novel approach for compensating the distortion of any interpolation based on modular method has been proposed. In this method the performance of the modular method is optimized by adding only some simply calculated coefficients. This approach causes drastic improvement in terms of signal-to-noise ratios with fewer modules compared to the classical modular method. Simulation results clearly confirm the improvement of the proposed method and also its superior robustness against additive noise.
\end{abstract}

\begin{IEEEkeywords}
Compensating distortion, interpolation, modular method, optimum coefficients.
\end{IEEEkeywords}

\section{Introduction}
\IEEEPARstart{D}{igital} to analog converters are common in digital signal processing and communication systems to reconstruct an analog signal from its discrete time samples. Several methods with different names were introduced in the literature in 1970's and 1980's \cite{lehmann}. S\&H and LI were the dominant methods before that time; today, Polynomial interpolation and B-Spline are the usual interpolation functions \cite{cubicConv, unser, cubicSpline}.

These interpolators create some distortion at the Nyquist rate after low pass filtering, especially when S\&H or LI are utilized. The advantage of these types of interpolators is their simplicity which makes them proper for practical use. To alleviate this problem, several methods such as inverse Sinc filtering, over-sampling, nonlinear and adaptive algorithms \cite{edge,bayesian,SAI}, a modular method of the recovery of a signal from its sampled-and-held version
are described in \cite{marvastiModular} for the uniform samples, \cite{Marvasti2} for the nonuniform samples, and successive approximation using an iterative method \cite{marvastiNonuniform2,marvastiIterative,marvastiNonuniform} were introduced. The modular method is compared to the inverse Sinc filtering in \cite{marvastiModular} which shows that by using a few numbers of modules, the performance of the modular method excels the inverse filtering as far as noise is concerned. Over-sampling is not a practical solution due to its bandwidth requirements. The iterative method \cite{marvastiIterative} outperforms the modular method at the cost of more computation.

We propose an optimized modular method which enhances the performance of the classical modular method \cite{ICT2011}. \footnote{Part of this paper is presented in 18th International Conference on Telecommunication (ICT) 2011.} Our method is based on some optimum coefficients which are computed very simply by solving a least square problem. Indeed, these coefficients are calculated just one time for a specific interpolation system and are independent of signal to which the modular method is going to be applied. The coefficients themselves do not increase the complexity of the modular method and is very simple for practical usages. The simulation results show that the coefficients are well optimized and perform better then classical method.

The rest of this paper is organized as follows: Section \ref{Preliminaries}, describes our general framework and introduces the terms and concepts used throughout the paper. Section \ref{Proposed} introduces our proposed method and is a straight forward manner to find the optimum coefficients for the modular method. Simulation results and comparison with the classical modular method for various interpolation systems will be presented in section \ref{Simulation} and finally, section \ref{Conlusion} will conclude this paper.

\section{Preliminaries}\label{Preliminaries}
In this section we give a brief overview of the modular method \cite{marvastiModular} that compensates the distortion of any interpolator such as Sample and Hold (S\&H) and linear order hold by mixing the sum of cosine waves and then passing them through a lowpass filter.
\IEEEpubidadjcol
Suppose $x(t)$ is sampled at the Nyquist rate ($\frac{1}{T}$) and assume $s(t)$ is any interpolating function that fits the samples of $x(t)$. According to these assumptions it can be formulated as follows:

\begin{equation}
s(t) = h(t)*\sum_{n=-\infty}^{+\infty}{x(nT)\delta(t-nT)}
\label{1}
\end{equation}
where $h(t)$ is the impulse response of the interpolation function. The above equation can be written as shown below in frequency domain:

\begin{equation}
S(f) = H(f)\times\sum_{i=-\infty}^{+\infty}{X(f-i/T)}
\label{2}
\end{equation}

According to the modular method, an improved reconstruction of $x(t)$ can be derived from $s(t)$ by following process:

\begin{equation}
\hat{x}(t) = s(t)\left(1+2cos(\frac{2\pi t}{T})+\dots+2cos(\frac{2N\pi t}{T})\right)*\Pi(t)
\label{3}
\end{equation}

where $\Pi(t)$ is a lowpass filter with a bandwidth equal to the bandwidth of $x(t)$ ($W$). Eq. (\ref{3}) can be rewritten in the frequency domain as follows:

\begin{eqnarray}
\hat{X}(f) &=& \Pi(f)\sum_{j=-N}^{+N}{S(f-j/T)}\nonumber\\
&=& \Pi(f)\sum_{j=-N}^{+N}{H(f-j/T)\sum_{i=-\infty}^{+\infty}{X(f-i/T-j/T)}}
\label{4}
\end{eqnarray}

Since $X(f)$  is band-limited (\ref{4}) can be simplified as follows:

\begin{equation}
\hat{X}(f) = X(f)\times\Pi(f)\sum_{j=-N}^{+N}{H(f-j/T)}
\end{equation}

\begin{figure}[t]
\centering
\includegraphics[width=90mm]{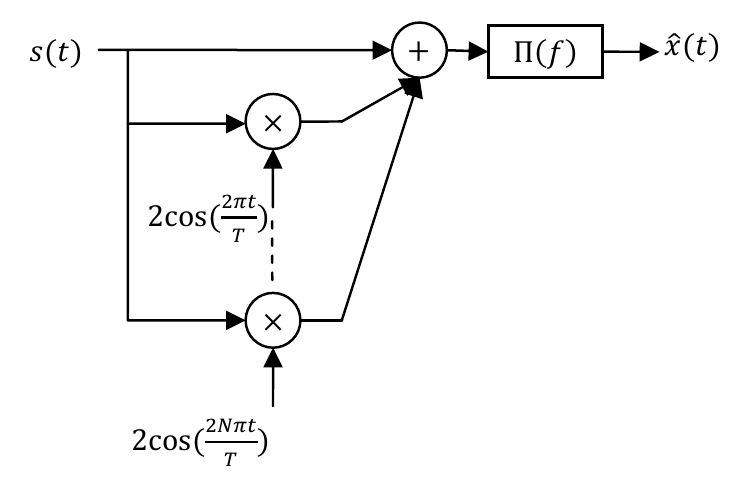}
\caption{The reconstruction block diagram using Modular Method}
\label{fig:modular}
\end{figure}

Therefore, it is obvious that as $N$ increases $\hat{x}$ will perfectly converge to $x$ if $\Pi(f)\sum_{j=-N}^{+N}{H(f-j/T)}$ becomes unity which always occurs for all interpolation functions since $\mathcal{F}^{-1}\{\sum_{j=-\infty}^{+\infty}{H(f-j/T)}\}=\delta(t)$, for instance this summation becomes summation of sinc functions which is unity for Sample and Hold (S\&H) interpolation function. However, practically it is not possible to apply infinite numbers of modules and only limited numbers of them are implemented by means of oscillators shown in Fig. \ref{fig:modular}. So the method will have distortion and to measure the distortion, let us define the mean-square error as:

\begin{equation}
e = \int_{-W}^{+W}{\left(\sum_{j=-N}^{+N}{H(f-j/T)}-1\right)^2}df
\label{5}
\end{equation}
and it is obvious that $\lim\limits_{N\rightarrow\infty}{e}=0$.

Another issue with this approach is that the signals are considered to be analog. In practice, most of the signals that we are dealing with are discrete especially in a computer for interpolation of images and audio files.

In the next section, we will drive all these relations again in the discrete domain and also minimize the mean square error using the optimization coefficients. The second part is the main part of this paper.

\section{Proposed Optimized Modular Method}\label{Proposed}
\subsection{Modular method in Diecrete Domain}
Consider $x[n]$ is band-limited discrete signal which is down sampled and interpolated at the Nyquist rate ($1/T$) by $h[n]$ and the result is $s[n]$. Therefore:

\begin{equation}
s[n] = h[n] * \sum_{k=-\infty}^{+\infty}{x[kT]\delta[n-kT]}
\label{6}
\end{equation}
and equivalently:

\begin{equation}
S(k) = H(k) * \sum_{i=-\infty}^{+\infty}{X((k-iN/T))_N}
\label{7}
\end{equation}
where $S(k)$, $H(k)$ and $X(k)$ respectively are N-point DFTs of $s[n]$, $h[n]$ and $x[n]$. Supposing that $N$ is divisible by $T$, the modular method can be formlated as follows:

\begin{equation}
\hat{x}[n] = \mbox{LPF}\big\{s[n]\left(1+2cos(2\pi n/T)+\dots+2cos(2M\pi n/T)\right)\big\}
\label{8}
\end{equation}
where LPF is a FFT lowpass filter. Also it can be shown that applying more than $[T/2]$ modules not only does not enhance the performance but also may distort it. Consider $T$ is even integer number and we have applied $k$ modules more than $T/2$, then we have:

\begin{eqnarray}
&1+\sum_{j=1}^{\frac{T}{2}+k}{2cos(\frac{2j\pi n}{T})}\nonumber\\
&=1+\sum_{ i=1}^{\frac{T}{2}}{2cos(\frac{2i\pi n}{T})}+\sum_{j=\frac{T}{2}+1}^{\frac{T}{2}+k}{2cos(\frac{2j\pi n}{T})}\nonumber\\
&=1+\sum_{i=1}^{\frac{T}{2}}{2cos(\frac{2i\pi n}{T})}+\sum_{j=1}^{k}{2cos(\frac{2(j+\frac{T}{2})\pi n}{T})}\nonumber\\
&=1+\sum_{i=1}^{\frac{T}{2}}{2cos(\frac{2i\pi n}{T})}+\sum_{i=1}^{k}{(-1)^n2cos(\frac{2i\pi n}{T})}
\end{eqnarray}
It is obvious that the third term in the above final result will distort previous modules; this effect can be shown for odd $T$s in the same way too. Hence, the maximum number of modules that is able to be applied is $[T/2]$.

Furthermore, in the case that the maximum number of modules are applied, if we put the multiplicand of last cosine $1$ instead of $2$, we will reach the impulse train. By means of Fourier series it would be proved as follows:
\begin{eqnarray}
\tilde{\delta}[n]&=&\sum_{i=-\infty}^{\infty}{\delta(n-iT)}=\sum_{T}{\frac{1}{T}e^{j2\pi kn/T}}\nonumber\\
&=&\frac{1}{T}\left[1+(-1)^n+\sum_{i=1}^{T/2-1}{2cos(2\pi in/T)}\right]\nonumber\\
&=&\frac{1}{T}\left[1+\sum_{i=1}^{T/2-1}{2cos(2\pi in/T)}+cos(2\pi \frac{T}{2}n/T)\right]
\end{eqnarray}
So, in this situation it will gather the original samples and if the filter is ideal, the output would be perfectly interpolated.\footnote{Also true for analog D/A where $2,2,\dots,2,1$ is not optimum but better than before.}

The equation (\ref{8}) will be rewritten in the frequency domain like below:

\begin{equation}
\hat{X}(k) = X(k)\times\Pi(k)\sum_{j=-M}^{+M}{H((k-jN/T))_N}
\label{9}
\end{equation}

Simultaneously, the error of the interpolation process would be formulated in the same manner performed in the pervious section:
\begin{equation}
e = \sum_{-N/T}^{+N/T}{\left(\sum_{j=-M}^{+M}{H((k-jN/T))_N}-1\right)^2}
\label{10}
\end{equation}

Again minimizing this error is our main goal and is related directly to the performance of our system. In next subsection, our proposed Technique will be introduced to achieve this purpose.
Again minimizing this error is our main goal and is related directly to the performance of our system. In next subsection, our proposed Technique will be introduced to achieve this purpose.

\subsection{Optimization Coefficients}
\begin{figure}[b]
\centering
\includegraphics[width=90mm]{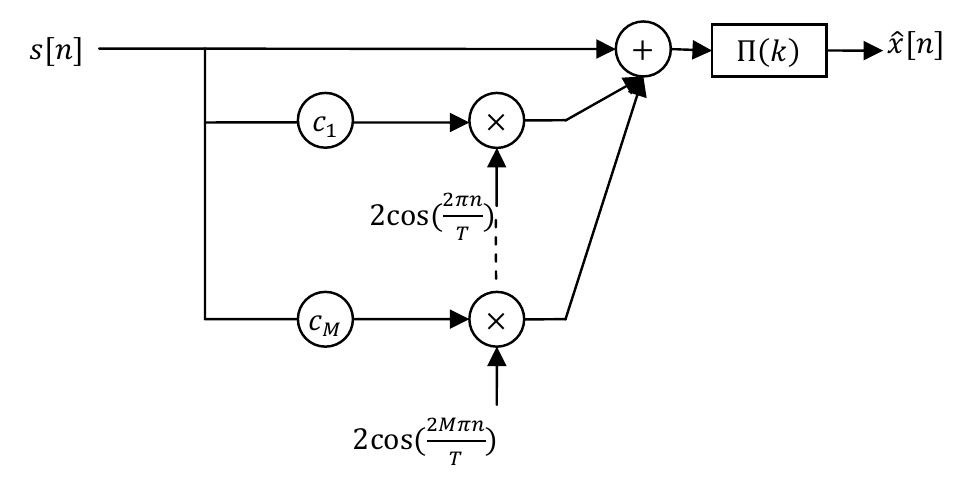}
\caption{The reconstruction block diagram using Discrete Modular Method and Optimazation coefficients}
\label{fig:Dmodular}
\end{figure}

Our goal is to minimize (\ref{10}) and as a result reduce distortion in the modular method in order to reach more precise interpolation for $x$. Our idea is that modules could be applied with some coefficients shown in Fig. \ref{fig:Dmodular}. By choosing these coefficients appropriately, the performance of the method increases efficiently and we can achieve the same result with fewer modules.

Now consider modules are multiplied by $c_j$:

\begin{equation}
\hat{x}[n] = \mbox{LPF}\Bigg\{s[n]\left(1+2c_1cos(\frac{2\pi n}{T})+\dots+2c_Mcos(\frac{2M\pi n}{T})\right)\Bigg\}
\label{11}
\end{equation}
and therefore, (\ref{10}) becomes:

\begin{equation}
e = \sum_{-N/T}^{+N/T}{\left(\sum_{j=-M}^{+M}c_{|j|}{H((k-jN/T))_N}-1\right)^2}
\label{12}
\end{equation}

For convenience the following parameters are defined:

\begin{eqnarray}
H_j(k)&\triangleq& H((k-iN/T))_N+H((k+iN/T))_N\nonumber\\
&=& \text{FFT}_N\{h[n]\times 2cos(\frac{2j\pi n}{T})\}
\label{13}
\end{eqnarray}

\begin{equation}
\mathfrak{H}=
\begin{pmatrix}
H_1(0) & H_2(0) & \ldots & H_M(0)\\
H_1(1) & H_2(1) & \ldots & H_M(1)\\
\vdots & \vdots & \ldots & \vdots\\
H_1(N/T) & H_2(N/T) & \ldots & H_M(N/T)\\
\end{pmatrix}
\label{14}
\end{equation}

\begin{equation}
\mathfrak{C}=
\begin{pmatrix}c_1\\c_2\\\vdots\\c_M\end{pmatrix},
\mathfrak{B}=
\begin{pmatrix}1\\1\\\vdots\\1\end{pmatrix}-\frac{1}{2}\begin{pmatrix}
H_0(0)\\H_0(1)\\\vdots\\H_0(N/T)\end{pmatrix}
\label{15}
\end{equation}

Now from these new definitions the error function can be simply rewritten as follows:

\begin{equation}
e = \left(\text{norm}(\mathfrak{H}\mathfrak{C}-\mathfrak{B})\right)^2
\label{16}
\end{equation}

Hence, to minimize this error, we should solve:

\begin{equation}
\mathfrak{H}\mathfrak{C}=\mathfrak{B}
\label{17}
\end{equation}

Since the number of equations ($N/T$) are much more than the unknowns ($M$), we could find the optimum answer by considering the mean square error method and the pseudo-inverse:

\begin{equation}
\mathfrak{C}=(\mathfrak{H}^T\mathfrak{H})^{-1}\mathfrak{H}^T\mathfrak{B}
\label{18}
\end{equation}

Therefore, the problem can be solved and we have found some coefficients which minimize the error for the finite number of modules without any exception on interpolation function. Furthermore, we will show in next section that these coefficients cause dramatic results in comparison to the classical modular method which is a special case of our method by assigning the coefficients one. The main key in our method is that the coefficients are calculated very easy and fast and by only accessing impulse response of interpolation function ($h[n]$). Moreover the coefficients are calculated just one time for an interpolation function and are stored in a lookup table and does not need to find them again every time we need them.

\subsection{Optimized Modular Method for 2-D Signals}
Due to increasing need to analyze the two-dimensional signals like image and video, we want to generalize our proposed method to 2-D space. For this reason, consider $x(t_1,t_2)$ is band-limited discrete signal which is sampled at the Nyquist rate ($\frac{1}{T_1} , \frac{1}{T_2}$) Now consider that this signal is reconstructed by interpolating function $h(t_1,t_2)$ and the resulted signal is $s(t_1,t_2)$. With no doubt the amount of signal $s(t_1,t_2)$ is equal to $x(t_1,t_2)$ in sampled points. By knowing these assumptions we can formulate them as follows:

\begin{equation}
s(t_1,t_2) = h(t_1,t_2) * \sum_{n_1,n_2=-\infty}^{+\infty}{x(n_1T_1,n_2T_2)\delta(t_1-n_1T_1,t_2-n_2T_2)}
\label{19}
\end{equation}

Easily the above equation, by a simple Fourier transform and by means of its properties in frequency domain, can be written like below:

\begin{equation}
S(f_1,f_2) = H(f_1,f_2)\times\sum_{i_1,i_2=-\infty}^{+\infty}{X(f_1-i_1/T_1,f_2-i_2/T_2)}
\label{20}
\end{equation}

By considering the modular method for one-dimensional signals we can understand that the main concept in modular method is rooted from averaging in frequency domain. Using this idea in one-dimensional domain we can generalize the modular method to two-dimensional domain. By knowing this, $2-D$ modular method should include the shifted version of that in different directions and here we use Lattice pattern. With these information, the reconstructed compensated signal $x(t_1,t_2)$ by means of $2-D$ modular method can be written from signal $s(t_1,t_2)$ is:

\begin{eqnarray}
\hat{x}(t_1,t_2) = s(x_1,x_2)(1+2cos(\frac{2\pi t_1}{T_1})+2cos(\frac{2\pi t_2}{T_2})\nonumber\\
+4cos(\frac{2\pi t_1}{T_1})+4cos(\frac{2\pi t_2}{T_2})+\dots)*\Pi(t)
\label{21}
\end{eqnarray}

Similar to one-dimensional signal we can easily transform the equation (\ref{21}) to the frequency domain:

\begin{eqnarray}
\hat{X}(f_1,f_2)&=& \Pi(f_1,f_2)\sum_{j_1,j_2=-N}^{+N}{S(f_1-j_1/T_1,f_2-j_2/T_2)}\nonumber\\
&=& \Pi(f_1,f_2)\sum_{j_1,j_2=-N}^{+N}{H(f_1-j_1/T_1,f_2-j_2/T_2)}\nonumber\\
&\times&
\sum_{i_1,i_2=-\infty}^{+\infty}{X(f_1-i_1/T_1-j_1/T_1,f_2-i_2/T_2-j_2/T_2)}\nonumber\\
\label{22}
\end{eqnarray}

We know that $X(f_1,f_2)$ is band limited signal. So, because of $\Pi(f_1,f_2)$\ in the equation (\ref{22}) that is a lowpass filter, the combination of signal $X(f_1,f_2)$ and its shifted one, the signal itself is resulted. Through the explanation, equation (\ref{22}) is simplified and can be written as follows:

\begin{equation}
\hat{X}(f_1,f_2) = X(f_1,f_2)\times\Pi(f_1,f_2)\sum_{j_1,j_2=-N}^{+N}{H(f_1-j_1/T_1,f_2-j_2/T_2)}
\end{equation}

Just like the idea used in one-dimensional method, when one condition exists, the equation perfectly converges to x when N increases toward $\infty$.

That condition is that the equation $\Pi(f_1,f_2)\sum_{j_1,j_2=-N}^{+N}{H(f_1-j_1/T_1,f_2-j_2/T_2)}$ becomes unity. By a little study, it is clear that this equation is true about all the interpolation functions, like the one that is proved in the one-dimensional method. Therefore, by increasing $N$, $\hat{x}$\ goes nearer to $x$.
But as mentioned in the last section, $N$ is the number of implementation modules that because of practical limitations on that, it cannot be increased that much. In the cases with limited $N$ the amount of the distortion from the main signal is the difference between 1 and $\Pi(f_1,f_2)\sum_{j_1,j_2=-N}^{+N}{H(f_1-j_1/T_1,f_2-j_2/T_2)}$. Then, we can formulate the error as the equation below:

\begin{equation}
e = \int_{-W_1}^{+W_1}\int_{-W_2}^{+W_2}{\left(\sum_{j_1,j_1=-N}^{+N}{H(f_1-j_1/T_1,f_2-j_2/T_2)}-1\right)^2}df_1df_2
\label{23}
\end{equation}

In this case, the only difference between the $2-D$ and $1-D$ is that $\mathfrak{C}$ and $\mathfrak{B}$ will be matrixes and the $\mathfrak{H}$ will be a tensor. With these definitions equation (21) can be written for two-dimensional case and it can be solved, so the compensated coefficients will be found.

\section{Simulation Results and Discussion}\label{Simulation}
We utilized MATLAB\textsuperscript{\textregistered} simulation environment to evaluate and compare the performance of methods. To have fair comparison, initial band limited signals are produced randomly, and the performance of each method is averaged over 100 signals. The initial signal is FFT lowpass filtered version of white Gaussian noise signals. To show the significance of this method, the sampling rate is performed at the Nyquist rate.

\begin{figure}[h]
\centering
\includegraphics[width=90mm]{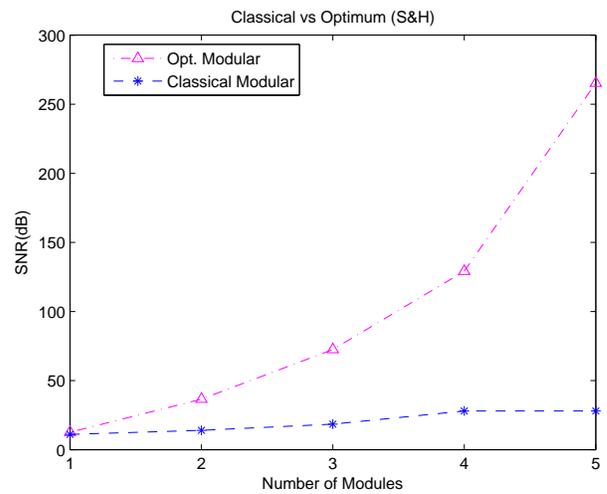}
\caption{SNR vs. the number of modules for Classical Modular method and our proposed Optimized method for S\&H interpolation.}
\label{fig:SH}
\end{figure}

The performance criterion for our simulations is the Signal to Noise Ratio (SNR) in dB. To avoid transient errors at the end points, SNR is calculated for interior points and 10\% of the end points are ignored. As illustrated in Fig. \ref{fig:SH}, the SNR increases monotonically in dB for classical method as the number of modules increases, while the optimum method increases exponentially as the number of modules increases. This means more than 250dB for simple S\&H interpolation and this is quite impressive in real engineering applications.

\begin{figure}[t]
\centering
\includegraphics[width=90mm]{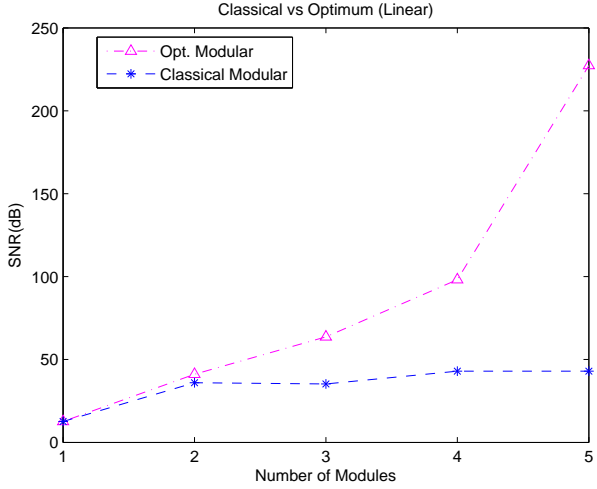}
\caption{SNR vs. the number of modules for Classical Modular method and our proposed Optimized Modular method for Linear interpolation}
\label{fig:Linear}
\end{figure}

Fig. \ref{fig:Linear} shows similar results for the Linear Interpolation (LI). The difference between the classical method and the optimum at the first few numbers of modules is not very significant. However, as the number of modules increases, the difference becomes apparent. This shows that our method can find the optimum coefficients for any interpolation independently.

\begin{figure}[b]
\centering
\includegraphics[width=90mm]{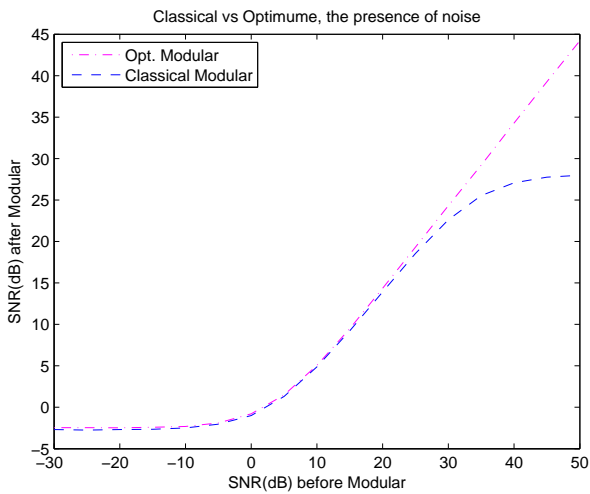}
\caption{SNR after applying methods vs. SNR of the signals before applyibg the Classical Modular method and our proposed Optimized method for  5 method and the S\&H interpolation.}
\label{fig:Noisy}
\end{figure}

To study the effect of noise, we added a white Gaussian noise to the band limited signal. This is the model of the electronic devices that generate thermal noise. Fig. \ref{fig:Noisy} shows that for low SNR, the performance of our method is not so significant over classical method, however as much as the power of noise decrease our method give much greater SNRs.

These simulations show that using the modular method by means of optimized coefficients enhance its performance dramatically both in noisy and noiseless environments independent of the type of interpolation.

\begin{figure}[t]
\centering
\includegraphics[width=90mm]{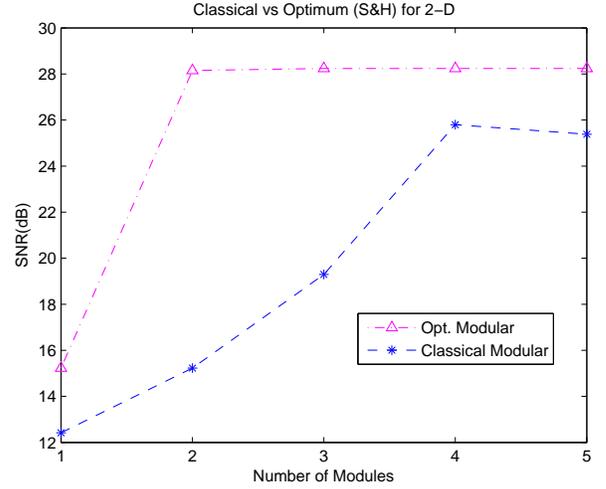}
\caption{SNR vs. the number of modules for Classical Modular method and our proposed Optimized method for S\&H interpolation for 2-D signals.}
\label{fig:SH2}
\end{figure}

Because of the importance of the two-dimensional signals, similar simulations are performed on them and their results are shown in Fig. \ref{fig:SH2} As is clear in figure, addition of coefficients has boosted the speed and with fewer modules we can reach better results than previous one. This shows inclusiveness of our method in other dimensions.

As in \cite{parandeh} and \cite{ICT2010}, by combination of two Modular and Iterative method we can reach significant results in interpolation. Also, here we tried to show the modular method's inclusiveness by means of compensated coefficients that are derived from this Hybrid Method. For this reason, we compared three methods, Classical Iterative (without modular), Hybrid Method (Iterative with Modular) and Hybrid Compensated Method (Iterative with Compensated Modular) with each other in both one-dimensional and two-dimensional domain versus number of iteration. As shown in figure \ref{fig:Iterative1} and \ref{fig:Iterative2} there is a significant progress in Iterative Method, so with just 10 iteration in 1-D and 13 iteration in 2-D we can reach its maximum possible value and this means saving time and hardware in exchange for calculation of coefficients that needs to be calculated just once.

Now to have a more tangible example for our method, we apply it to image enlargements. For this purpose we use Lena's known image. First after a continuous sampling the $512\times512$ image is converted to $256\times256$ image. Then, this small image is going to be upsampled by means of various and up to date methods. Results are compared to the main image by performance criterion PSNR in dB. The results are written in the table \ref{tab:1}. According to these results our method in iterative case has better performance than all the existing methods, while it has less complexity in comparison with many of
these methods.

For a visual comparison, the upsampled images by various methods are shown in figure 8. With a little precision, these differences are apparent in the image, specially, in the edge of the hat and eyes the difference is obvious.
All the simulations show the capability of the recommended method in various fields.

\section{Conclusion}\label{Conlusion}
A novel Optimized Modular method is proposed for compensating error of any interpolation system. We add the optimized coefficients calculated in a very simple manner into the Classical Modular method in order to maximize its performance as much as it could be. Also, the proposed method is generalized to two-dimensional domain. The simulations show

\onecolumn
\begin{figure}[p]
\centering
\subfigure[]
{\label{fig:Original}\includegraphics[width=70mm]{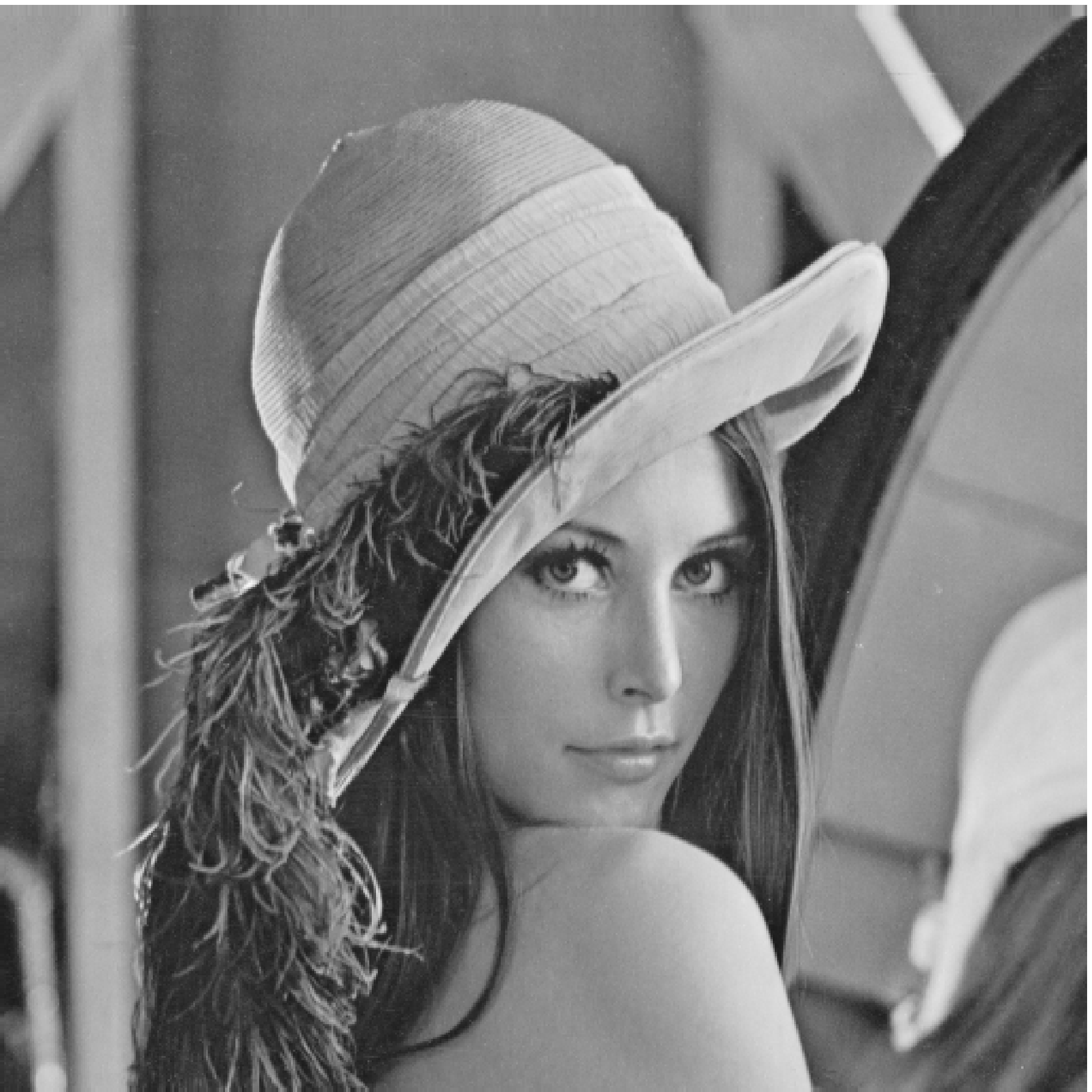}}%
~~~~~\subfigure[]
{\label{fig:Bilinear}\includegraphics[width=70mm]{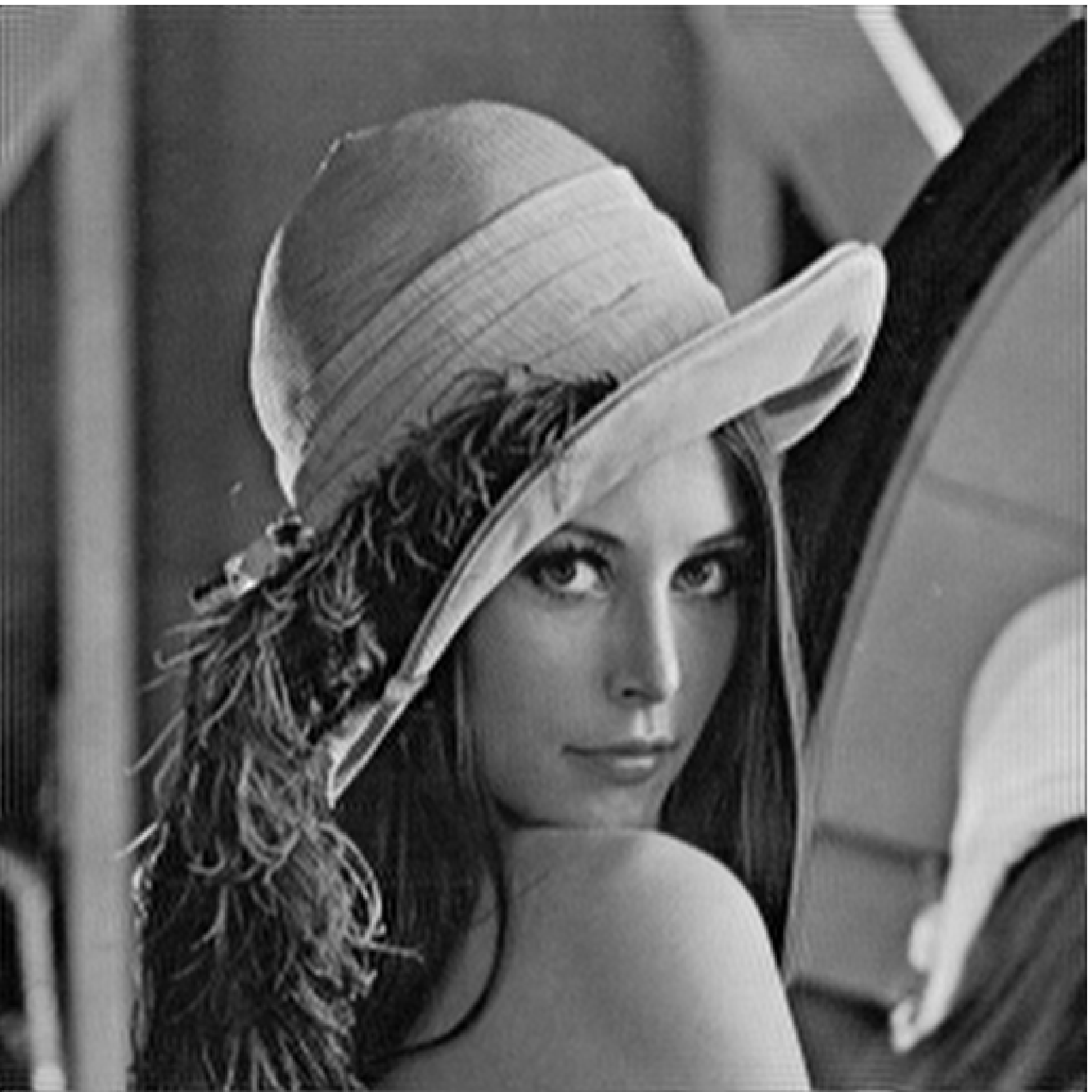}}
\subfigure[]
{\label{fig:Bicubic}\includegraphics[width=70mm]{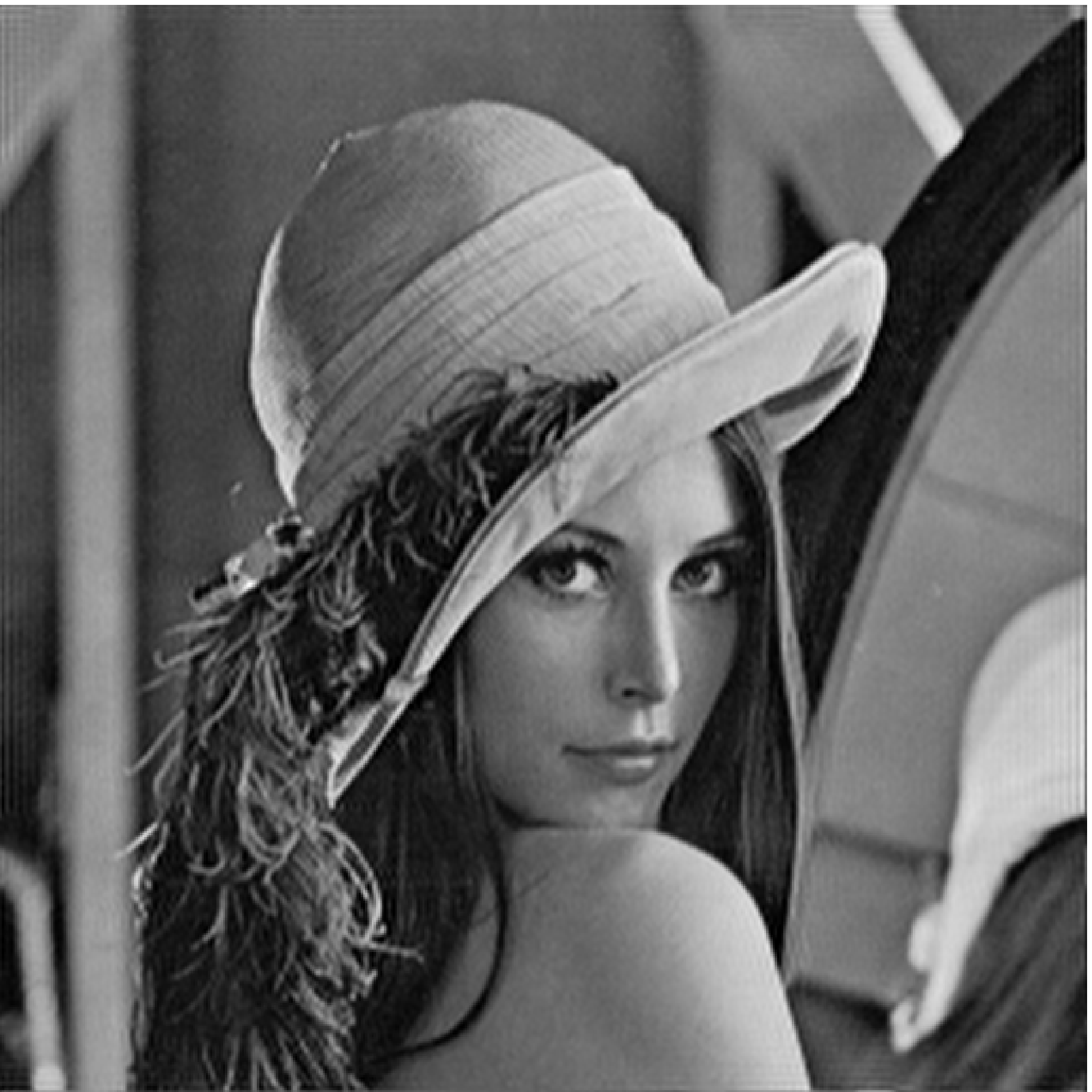}}%
~~~~~\subfigure[]
{\label{fig:WZPCS}\includegraphics[width=70mm]{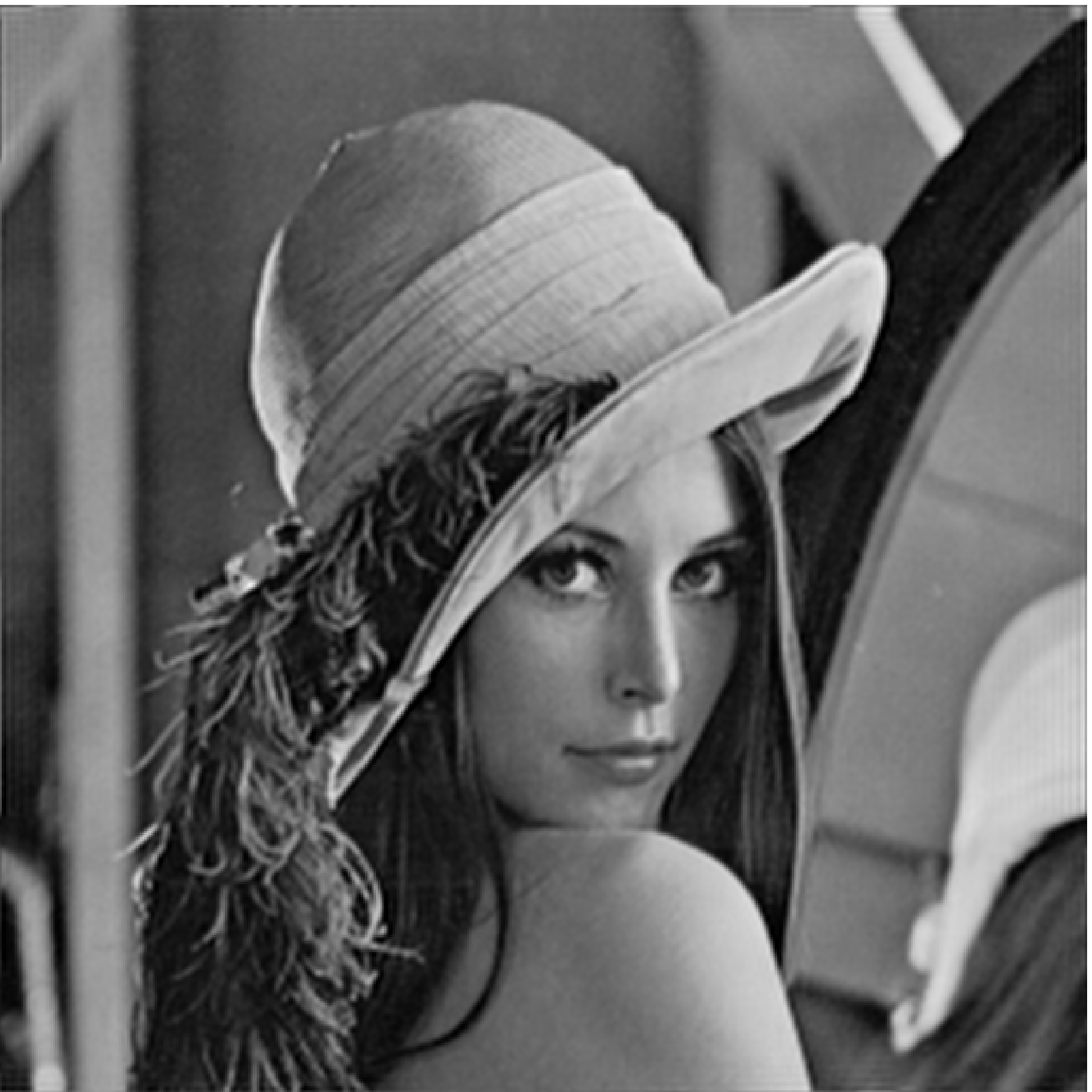}}
\subfigure[]
{\label{fig:SAI}\includegraphics[width=70mm]{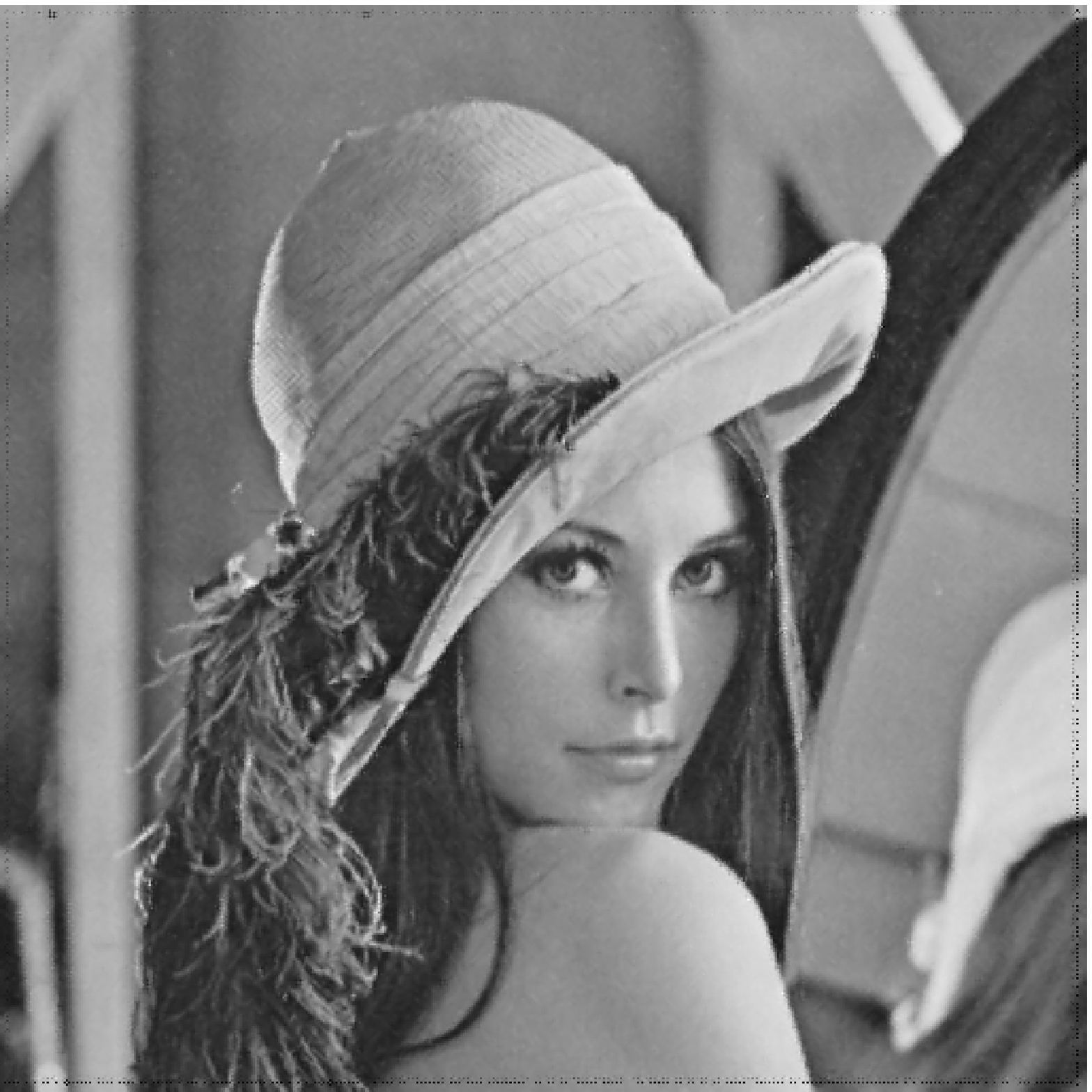}}%
~~~~~\subfigure[]
{\label{fig:Opt}\includegraphics[width=70mm]{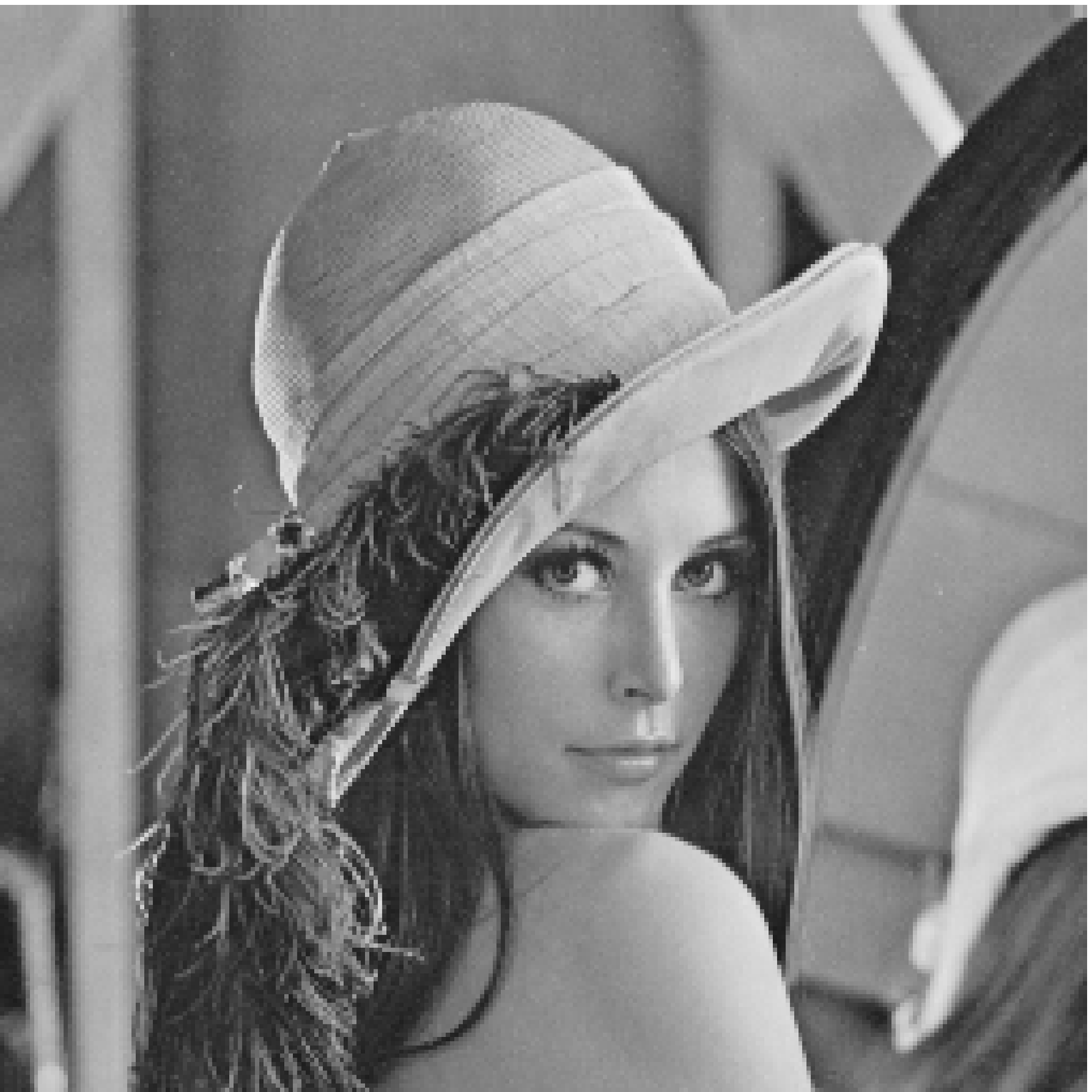}}
\caption{Comparing various methods, (a) main image, (b) Optimized Hybrid, (c) Hybrid, (d) Iterative, (e) SAI and (f) WZP}
\label{fig:Lena}
\end{figure}

\twocolumn

\begin{figure}[h]
\centering
\subfigure[1-D]
{\label{fig:Iterative1}\includegraphics[width=90mm]{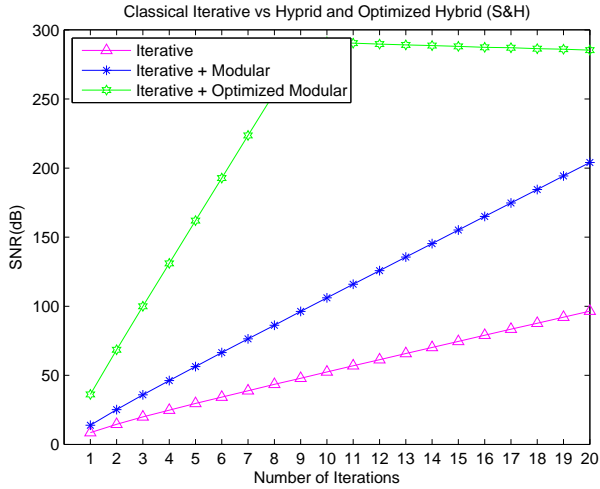}}
\subfigure[2-D]
{\label{fig:Iterative2}\includegraphics[width=90mm]{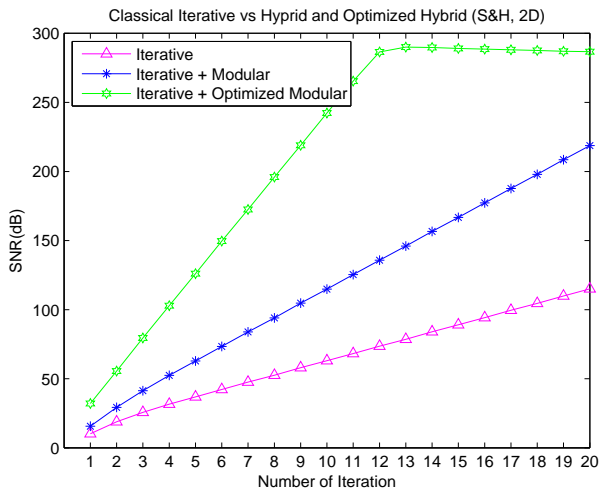}}
\caption{Comparison of 3 method, Classical Iterative, Hybrid Method and Hybrid Compensated Method vs. iteration number for S\&H in (a)1-D and (b)2-D}
\end{figure}

very high improvements versus classical methods. This proposed method is also more favorable in terms of computational complexity with respects to other error compensating methods, since not only the modular method has very simple algorithm but also we could reach better SNR values with just 2 modules rather than 5 modules in the classical method. In addition, combining this method with Iterative Method can result in a more powerful method and has better performance in both one-dimensional and two-dimensional domain in comparison with many other interpolation methods exist in world.

\begin{table}[t]
\begin{center}
\caption{COMPARISON OF THE RESULTS OF LENA'S IMAGE BY VARIOUS METHODS VS. PSNR (DB)}
\label{tab:1}
\begin{tabular}{|c|c|}
\hline
\textbf{Method}&\textbf{Lena}\\\hline\hline
Bilinear&30.13\\\hline
Bicubic&31.34\\\hline
NEDI\cite{temizel}&34.10\\\hline
WZP-Haar\cite{temizel}&31.46\\\hline
WZP-Db.9/7\cite{temizel}&34.45\\\hline
Carey et al.\cite{Carey}&34.48\\\hline
HMM\cite{kinebuchi}&34.52\\\hline
HMM-SR\cite{zhao}&34.61\\\hline
WZP-CS\cite{temizel}&34.93\\\hline
SAI\cite{SAI}&34.74\\\hline
Iterative (2 iter.)\cite{marvastiIterative}&35.25\\\hline
Iterative (10 iter.)\cite{marvastiIterative}&37.39\\\hline
Hybrid (2 iter. and 1 mod.)\cite{ICT2010}&37.12\\\hline
\textbf{Opt. Hybrid (2 iter. and 1 mod.)}&\textbf{37.41}\\\hline
\end{tabular}
\end{center}
\end{table}

\nocite{*}
\bibliographystyle{IEEEtran}
\bibliography{meee}

\end{document}